# Probabilistic Models for Agents' Beliefs and Decisions

**Brian Milch**
Computer Science Department
Stanford University
Stanford, CA 94305-9010
*milch@cs.stanford.edu*

**Daphne Koller**
Computer Science Department
Stanford University
Stanford, CA 94305-9010
*koller@cs.stanford.edu*

## Abstract

Many applications of intelligent systems require reasoning about the mental states of agents in the domain. We may want to reason about an agent's beliefs, including beliefs about other agents; we may also want to reason about an agent's preferences, and how his beliefs and preferences relate to his behavior. We define a probabilistic epistemic logic (PEL) in which belief statements are given a formal semantics, and provide an algorithm for asserting and querying PEL formulas in Bayesian networks. We then show how to reason about an agent's behavior by modeling his decision process as an influence diagram and assuming that he behaves rationally. PEL can then be used for reasoning from an agent's observed actions to conclusions about other aspects of the domain, including unobserved domain variables and the agent's mental states.

## 1 Introduction

When an intelligent system interacts with other agents, it frequently needs to reason about these agents' beliefs and decision-making processes. Examples of systems that must perform this kind of reasoning (at least implicitly) include automated e-commerce agents, natural language dialogue systems, intelligent user interfaces, and expert systems for such domains as international relations. A central problem in many domains is predicting what other agents will do in the future. Since an agent's decisions are based on its beliefs and preferences, reasoning about mental states is essential to making such predictions. An equally important task is making inferences about the state of the world based on another agent's beliefs (possibly revealed through communication) and decisions. Since other agents often observe variables that are hidden from our intelligent system, their beliefs and decisions may provide information about the world that the system cannot obtain by other means.

Suppose, for example, that we are developing a system to help analysts and policymakers reason about international crises. In one example, based loosely on a scenario presented in [3], Iraq purchases weapons-grade anthrax (a deadly bacterium) and begins to develop a missile capable of delivering anthrax to targets in the Middle East. There is a vaccine against anthrax which the United States is currently administering to its troops, but for ethical reasons the U.S. has not done controlled studies of the vaccine's effectiveness. Iraq, on the other hand, may have performed such tests. Iraq's purpose in attempting to develop an anthrax-equipped missile is to strike U.S. Air Force personnel in Turkey or Saudi Arabia, inflicting as many casualties as possible. However, if Iraq works on developing the missile, it must use an old weapons plant that is prone to fire; a fire at the plant would be visible to U.S. satellites. We would like our intelligent system to be able to answer questions like, "If we observe that Iraq has purchased anthrax, what is the probability that the vaccine is effective?", and "Does Iraq believe (e.g., with probability at least 0.3) that if they begin developing an anthrax-carrying missile, the U.S. will realize (e.g., believe with probability at least 0.9) that they have acquired anthrax?".

Efforts to formalize reasoning about beliefs date back to Hintikka's work on epistemic logic [6]. The classical form of epistemic logic does not allow us to quantify an agent's uncertainty about a formula; we can only say that an agent knows $\varphi$ or does not know $\varphi$. Probabilistic logics of knowledge and belief [4, 16] remove this limitation. However, evaluating the probability that an agent $a$ assigns to a formula $\varphi$ in a model of one of these logics requires evaluating $\varphi$ at every state that $a$ considers possible. As the number of states is exponential in the number of domain variables, this process is computationally intractable.

One of the main contributions of this paper is the introduction of a *probabilistic epistemic logic (PEL)* that uses Bayesian networks (BNs) [12] as a compact representation for the agents' beliefs. This framework allows us to perform probabilistic epistemic inference without enumerating an exponential number of states. Our approach is based on the **common prior assumption** common in economics [1]. It states that the agents have a common prior probability distribution over outcomes and their beliefs differ only because they have different observations; this assumption allows us to use a single BN for representing all the agents' beliefs. We describe an implemented algorithm for adding nodes to this BN so that it can be used to evaluate arbitrary PEL formulas.

In most domains, agents do not just passively observe the world and form beliefs; they also make decisions and act. In many existing probabilistic reasoning systems (e.g., [2, 13, 7]), a human expert defines the conditional probability distributions (CPDs) that describe how likely an agent is to take each possible action, given an instantiation of the variables relevant to the agent's decision process. But this technique relies on a human's understanding of how agents make decisions, and it may be difficult for a human to perform such analysis for a complex model. If we assume an agent acts rationally, the intelligent system can use decision theory to derive the CPDs for the agent's actions automatically. This problem involves subtle strategic (game-theoretic) reasoning when multiple agents are acting and have uncertainty about each other's actions [5]. In this paper we restrict attention to the case where only one agent acts. We model the agent's decision process using an influence diagram (ID) [8], then convert this influence diagram into a Bayesian network. This extension allows us to use PEL in order to reason about the decision-maker's likely course of action, and (more interestingly) to use his actions to reach conclusions about unobserved aspects of the world. We can also extend the framework to reach conclusions about the decision-maker's preferences, which may not be common knowledge.

## 2 A Probabilistic Epistemic Logic

Our probabilistic epistemic logic (PEL) is essentially a special case of the logic of knowledge and belief defined by Fagin and Halpern [4] (FH hereafter). In PEL, we assume that agents have a common prior probability distribution over states of the world, and an agent's local probability distribution at state $s$ is equal to this global distribution conditioned on the set of states the agent considers possible at $s$. These assumptions are not uncontroversial, but we will defer a discussion of the alternatives until Section 6.

The language of PEL is parameterized by a set $\Phi$ of random variable symbols, each with an associated domain; a set $\mathcal{A}$ of agents; and a number $N_a$ of observation stages for each agent $a \in \mathcal{A}$. At each of an agent's observation stages, there is a certain set of variables whose values the agent has observed. In this paper, we will make the **perfect recall assumption**: agents do not forget observations they have made. The values of the variables themselves do not change from stage to stage (if we want to model an aspect of the world that changes over time, we must create separate variables for separate times).

Given these parameters, the language of PEL consists of the following:

- *atomic formulas* of the form $X = v$, where $X \in \Phi$ and $v \in dom(X)$ (the domain of $X$). Note that $dom(X)$ need not be $\{true, false\}$; it may be any non-empty finite set.
- formulas of the form $\neg \varphi$ and $\varphi \vee \psi$, where $\varphi$ and $\psi$ are PEL formulas; we use $\varphi \wedge \psi$ as an abbreviation for $\neg(\neg \varphi \vee \neg \psi)$.
- formulas of the form $BelCond_{a,i}^{\geq r}(\varphi \mid \psi)$, where $a \in \mathcal{A}$, $i \in \{1, \ldots, N_a\}$, $\varphi$ and $\psi$ are PEL formulas, and $r$ is a probability in $[0, 1]$.

Our atomic formulas play the same role as propositions in the FH logic. The modal formula $BelCond_{a,i}^{\geq r}(\varphi \mid \psi)$ should be read as, "according to agent $a$ in stage $i$, the conditional probability of $\varphi$ given $\psi$ is at least $r$". The unconditional belief operator $Bel_{a,i}^{\geq r}(\varphi)$ is an abbreviation for $BelCond_{a,i}^{\geq r}(\varphi \mid true)$. We will provide formal semantics for these statements after defining a model theory for PEL. Note that the ability to express conditional belief statements is not included in the FH logic, although their belief statements are more expressive than ours in allowing probabilities to be related by arbitrary linear inequalities.

**Definition 1** *A model $M$ of the PEL language having random variables $\Phi$, agents $\mathcal{A}$ and observation process lengths $\{N_a\}_{a \in \mathcal{A}}$ is a tuple $(S, \pi, \mathcal{K}, \mathcal{P})$, where:*

- *$S$ is a set of possible states of the world;*
- *$\pi$ is a value function mapping each random variable symbol $X \in \Phi$ to a discrete random variable $X^M$ (a function from $S$ to $dom(X)$);*
- *$\mathcal{K}$ maps each pair in $\{(a, i) \in \mathcal{A} \times \mathbb{Z}^+ : i \leq N_a\}$ to an accessibility relation $\mathcal{K}_{a,i}$ which is an equivalence relation on $S$;*
- *$\mathcal{P}$ is a probability distribution over $S$.*

Thus, a PEL model specifies a set of states and maps each random variable symbol to a random variable defined on those states. In the rest of the paper, we will often refer to a random variable $X^M$ simply as $X$; it should be clear from context whether a random variable or a random variable symbol is intended. The accessibility relation $\mathcal{K}_{a,i}$ holds between worlds that are indistinguishable to agent $a$ at stage $i$. In other words, at stage $i$, if $s$ and $s'$ are in the same accessibility equivalence class, agent $a$ has no information that allows him to distinguish between world $s$ and world $s'$. We will use the notation $\mathcal{K}_{a,i}(s)$ to refer to the set of states $s' \in S$ such that $\mathcal{K}_{a,i}(s, s')$. With this semantics, the perfect recall assumption is formalized as a requirement that if $\neg \mathcal{K}_{a,i}(s, s')$, then for all $j > i$, we also have that $\neg \mathcal{K}_{a,j}(s, s')$.

$\mathcal{P}$ is the agents' common prior probability distribution over the set of states $S$. For each agent $a$, stage $i$, and state $s$, we can derive a local distribution $p_{a,i,s}$ over the states accessible from $s$. This local distribution is the subjective probability that the agent assigns to each accessible state.

**Definition 2** *Consider any $a \in \mathcal{A}$, $i \in \{1, \ldots, N_a\}$, and $s \in S$. Then for each state $s' \in \mathcal{K}_{a,i}(s)$, we define: $p_{a,i,s}(s') = \mathcal{P}(s' \mid \mathcal{K}_{a,i}(s))$.*

Note that an agent $a$'s subjective probability distribution varies from state to state. Thus, other agents' uncertainty about the state of the world can lead to uncertainty about $a$'s beliefs. For example, in some states Iraq believes the anthrax vaccine to be effective, and in other states it does not; the U.S. may not be able to distinguish these two kinds of states.

The semantics of PEL will be familiar to readers with background in modal logic. We introduce a satisfaction relation $\models$, such that $(M, s) \models \varphi$ means the formula $\varphi$ is satisfied at world $s$ in model $M$. We also define an inverse relation $[\varphi]_M = \{s \in S : s \models \varphi\}$.

**Definition 3** $(M, s) \models \varphi$ *if one of the following holds:*
- $\varphi$ is an atomic formula $X = v$ and $X(s) = v$.
- $\varphi = \neg \psi$ and $(M, s) \not\models \psi$.
- $\varphi = \psi \vee \chi$, and $(M, s) \models \psi$ or $(M, s) \models \chi$.
- $\varphi = BelCond_{a,i}^{\geq r}(\psi \mid \chi)$, $p_{a,i,s}([\chi]_M) \neq 0$, and $p_{a,i,s}([\psi]_M \mid [\chi]_M) \geq r$.

Note that if there are no states accessible from $s$ that satisfy $\chi$, then $BelCond_{a,i}$ is defined to be false.

This definition of satisfaction allows us to evaluate a PEL formula at any state $s$ in a given model $M$. We can then use the prior probability distribution $\mathcal{P}$ to find the total probability of states that satisfy a formula $\varphi$. If we do this evaluation directly in the PEL model, we need to evaluate $\varphi$ at each of $|S|$ states, and the size of the state space can be quite large — typically exponential in the number of variables. In the next section, we present a representation for PEL models based on Bayesian networks, and an algorithm that uses the independence assumptions encoded in the BN to find the probabilities of PEL formulas efficiently.

Thus, we are proposing an efficient model-checking procedure for PEL formulas. We could also provide a proof system for PEL; in fact, Fagin and Halpern provide a complete axiomatization for their logic. However, it is reasonable to assume that an intelligent agent will have a complete model representing its own belief state, and it is often more efficient to assert and query formulas in a model than to attempt to derive formulas from a knowledge base (which would need to be quite large to completely define the agent's beliefs).

## 3 Representing a PEL model as a BN

Bayesian networks provide a compact representation of a complex probability space. We can augment Bayesian networks to provide a compact representation of a certain class of PEL models. The basic idea is as follows. We define a PEL model $M$ over the set of random variables $\Phi$ using a BN $\mathcal{B}$ that has a node for each $X \in \pi(\Phi)$. We define $S$ to be the set of all possible assignments $\mathbf{x}$ to the variables in $\pi(\Phi)$. The distribution defined by $\mathcal{B}$ specifies the distribution $\mathcal{P}$ over $S$.

To define the accessibility relation $\mathcal{K}_{a,i}$ in this framework, we place the restriction that an agent's observations always correspond to some set of random variables:

**Observation Set Assumption:** For every $a \in \mathcal{A}$ and $i \in \{1, \ldots, N_a\}$, there is an *observation set* $\mathcal{O}_{a,i} \subset \pi(\Phi)$ such that:
$$\mathcal{K}_{a,i}(s, s') \iff \forall X \in \mathcal{O}_{a,i} \ (X(s) = X(s'))$$

Given this assumption, the perfect recall assumption is equivalent to the requirement that if $j > i$, then $\mathcal{O}_{a,i} \subseteq \mathcal{O}_{a,j}$.

**Definition 4** *Let $M = (S, \pi, \mathcal{K}, \mathcal{P})$ be a PEL model; let $\mathcal{B}$ be a BN defining a joint distribution $\Pr$ and let $\mathcal{O}_{a,i}$ be observation sets consisting of random variables appearing in $\mathcal{B}$. We say that $M$ and $\langle \mathcal{B}, \{\mathcal{O}_{a,i}\} \rangle$ are equivalent if:*
- *for every $X \in \Phi$, $X$ is in $\mathcal{B}$;*
- *for any instantiation $\mathbf{x}$ of $\pi(\Phi)$, $\mathcal{P}(\mathbf{x}) = \Pr(\mathbf{x})$;*
- *for each agent $a$ and stage $i$, $\mathcal{K}_{a,i}$ is related to $\mathcal{O}_{a,i}$ according to the Observation Set assumption.*

We can now use this framework to model the scenario described in the introduction. The equivalent Bayesian network is shown in Figure 1. Let $i$ stand for Iraq and $u$ stand for the United States. We assume that Iraq has a six-stage observation process: $\mathcal{O}_{i,1} = \{V\}$; $\mathcal{O}_{i,2} = \{V, P\}$; $\mathcal{O}_{i,3} = \{V, P, B\}$;

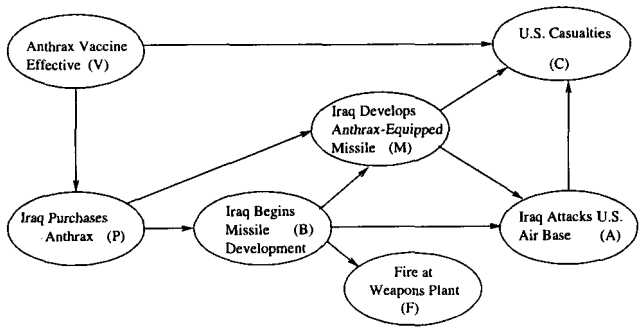

Figure 1: Basic Bayesian network for the crisis management problem.

$\mathcal{O}_{i,4} = \{V, P, B, F, M\}$; $\mathcal{O}_{i,5} = \{V, P, B, F, M, A\}$; and $\mathcal{O}_{i,6} = \{V, P, B, F, M, A, C\}$. Meanwhile, the U.S. has $\mathcal{O}_{u,1} = \emptyset$; $\mathcal{O}_{u,2} = \{F\}$; and $\mathcal{O}_{u,3} = \{F, A, C\}$.

Before it decides whether to attack the U.S. air base, is Iraq quite sure that U.S. casualties will be either high or medium? We can answer this question by evaluating the formula $Bel_{i,4}^{\geq 0.8}((C = high) \vee (C = medium))$. A more complex query is "Does Iraq believe with probability at least 0.3 that if they begin developing an anthrax-carrying missile, the U.S. will believe with probability at least 0.9 that they have acquired anthrax?". If we fill in the stages of the observation processes that are implicit in this question, we get the PEL formula $BelCond_{i,2}^{\geq 0.3} \left( Bel_{u,2}^{\geq 0.9} (P = true) \mid B = true \right)$.

The Observation Set Assumption implies that it is common knowledge what variables agent $a$ has observed at stage $i$. In many cases, this assumption is unrealistic; in our example, the U.S. might be uncertain whether Iraq observed the effectiveness of the anthrax vaccine at stage 1. As we show, we can handle such situations without modifying PEL. We simply add a new node $Observes_{i,1}(V)$ to the BN of Figure 1. This node is $true$ if Iraq has observed $V$ at stage 1, and $false$ otherwise; it can have as parents any nodes that are not descendents of $V$. We also add a node $ObservedValue_{i,1}(V)$ that has $V$ and $Observes_{i,1}(V)$ as parents. Its domain is $dom(V) \cup \{unknown\}$. It takes the value $unknown$ if $Observes_{i,1}(V)$ is $false$, but has the same value as $V$ if $Observes_{i,1}(V)$ is $true$. We let $\mathcal{O}_{i,1}$ contain $Observes_{i,1}(V)$ and $ObservedValue_{i,1}(V)$, but not $V$ itself.

Under this construction, it is common knowledge that Iraq knows at stage 1 whether it has observed $V$ at stage 1, and knows what value it has observed. However, since the value of $Observes_{i,1}(V)$ is not common knowledge, the U.S. may not know whether $ObservedValue_{i,1}(V)$ has the uninformative $unknown$ value, or is a copy of $V$. We have defined this construction process with an example, but it is clearly general enough to model uncertainty about whether any variable $X$ is in any observation set $\mathcal{O}_{a,i}$. The modified BN and observation sets now define a PEL model over a richer set of states than simply the possible instantiations of $\pi(\Phi)$.

## 4 Evaluating PEL Formulas in a BN

This framework allows us to represent a PEL model compactly, but how do we answer queries such as the ones shown above? We can use an equivalent BN $\mathcal{B}$ to find the probability $\mathcal{P}(X = v)$ of any atomic formula, simply by finding $\Pr(X = v)$. We want to extend $\mathcal{B}$ so that it allows us to compute the probability of an arbitrary PEL formula $\varphi$. To this end, we first define an indicator variable $\eta[\varphi]$ which is $true$ if $M, s \models \varphi$ and $false$ otherwise. We then extend the BN to include not only the random variables $\pi(\Phi)$, but also indicator variables for some set $\Delta$ of formulas that we may assert or query. Since both $\pi(\Phi)$ and all such indicator variables are defined on $S$, the distribution $\mathcal{P}$ over $S$ defines a joint distribution for $\pi(\Phi) \cup \eta[\Delta]$. Our goal in constructing the augmented BN is to ensure that it defines the same joint distribution.

**Definition 5** Let $M = (S, \pi, \mathcal{K}, \mathcal{P})$ be a PEL model; let $\mathcal{B}$ be an augmented BN defining a joint distribution $\Pr$ and let $\mathcal{O}_{a,i}$ be observation sets. Let $\Delta$ be a set of PEL formulas. Then $M$ and $\langle \mathcal{B}, \{\mathcal{O}_{a,i}\}\rangle$ are $\Delta$-equivalent if $M$ and $\mathcal{B}$ are equivalent and:
- for every $\varphi \in \Delta$, $\eta[\varphi]$ is in $\mathcal{B}$;
- for any instantiation $\boldsymbol{x}$ of $(\pi(\Phi) \cup \eta[\Delta])$, $\mathcal{P}(\boldsymbol{x}) = \Pr(\boldsymbol{x})$.

We now present an algorithm that, given a BN that is equivalent to a PEL model $M$, adds indicator variables to create an augmented BN that is $\Delta$-equivalent to $M$, for an arbitrary set of formulas $\Delta$. The central function of our algorithm is **CreateNode**$(\mathcal{B}, \varphi)$, which takes as arguments a BN $\mathcal{B}$ and a PEL formula $\varphi$. Its purpose is to create an indicator node for $\varphi$, store it in a global table, and give it the proper conditional distribution given the other variables in $\mathcal{B}$.

If there is already a node $\eta[\varphi]$ in the table, **CreateNode** returns immediately. If $\varphi$ is an atomic formula $X = v$, then the function creates a node $\eta[\varphi]$ whose sole parent is $X$. It defines the CPD of $\eta[\varphi]$ such that $\eta[\varphi] = true$ (with probability 1) if $X = v$, and $\eta[\varphi] = false$ otherwise.

If $\varphi = \neg \psi$, the function calls **CreateNode**$(\mathcal{B}, \psi)$. Then it creates a node $\eta[\varphi]$ with one parent, $\eta[\psi]$. It defines the CPD of $\eta[\varphi]$ like a NOT gate: $\eta[\varphi] = true$ iff $\eta[\psi] = false$. If $\varphi = \psi \vee \chi$, the function calls

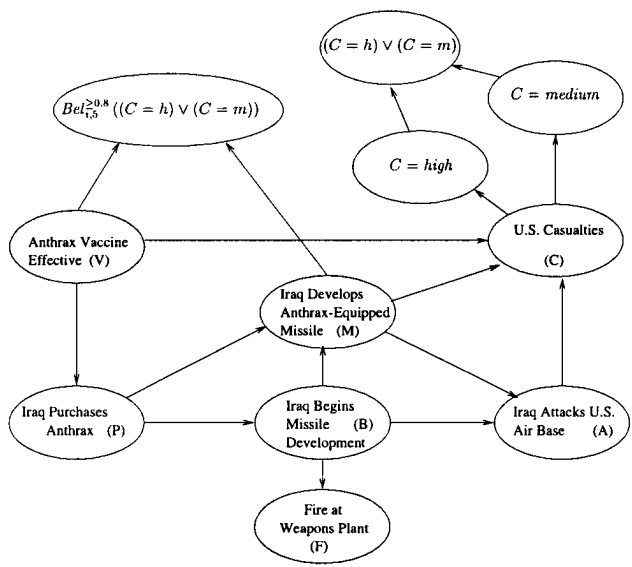

Figure 2: Bayes net with indicator variables added.

**CreateNode**$(\mathcal{B}, \psi)$ and **CreateNode**$(\mathcal{B}, \chi)$. Then it creates a node $\eta[\varphi]$ with two parents, $\eta[\psi]$ and $\eta[\chi]$. In this case, the CPD for $\eta[\varphi]$ is like an OR gate: $\eta[\varphi] = true$ iff $\eta[\psi] = true$ or $\eta[\chi] = true$.

The interesting case is where $\varphi = BelCond_{a,i}^{\geq r}(\psi \mid \chi)$. As usual, the function begins by calling **CreateNode**$(\mathcal{B}, \psi)$ and **CreateNode**$(\mathcal{B}, \chi)$. Now, recall that $\mathcal{O}_{a,i}$ is the set of variables whose values agent $a$ has observed at stage $i$. Clearly, whether the agent assigns probability at least $r$ to $\psi$ given $\chi$ depends on what the agent has observed. However, it may be that not all the observations are relevant; some of the variables in $\mathcal{O}_{a,i}$ may be d-separated from $\eta[\psi]$ given the other observations and $\eta[\chi]$. Using an algorithm such as that of [14], **CreateNode** determines the minimal subset **Rel** $\subset \mathcal{O}_{a,i}$ of *relevant observations* such that $\mathcal{O}_{a,i}$ − **Rel** is d-separated from $\eta[\psi]$ given **Rel** $\cup \{\eta[\chi]\}$. It then creates a node $\eta[\varphi]$ with the elements of **Rel** as parents. Next, **CreateNode** sets $\eta[\chi] = true$ as evidence, and uses a BN inference algorithm (e.g., [10]) to obtain a joint distribution over $\eta[\psi]$ and **Rel**. For each instantiation **rel** of **Rel**, the function uses the joint distribution to calculate $\Pr(\eta[\psi] \mid \langle \mathbf{rel}; \eta[\chi] = true \rangle)$. We then set the CPD $P(\eta[\varphi] \mid \mathbf{rel})$ to give probability 1 to *true* if $\Pr(\psi \mid \langle \mathbf{rel}; \eta[\chi] = true \rangle) \geq r$ and probability 1 to *false* otherwise.

As an example of how this algorithm works, consider the formula we discussed earlier involving Iraq's beliefs about U.S. casualties:

$$\varphi = Bel_{i,4}^{\geq 0.8}((C = high) \vee (C = medium))$$

Calling **CreateNode** on this formula results in a recursive call to create a node for $((C = h) \vee (C = m))$, which in turn calls **CreateNode** for $(C = h)$ and $(C = m)$. There are five random variables in Iraq's observation set at stage 4, but it turns out that only two, $V$ (vaccine effective) and $M$ (missile developed), are relevant to $((C = h) \vee (C = m))$. To obtain the CPD for $\eta[\varphi]$, we perform BN inference to calculate $\Pr(\eta[(C = h) \vee (C = m)] \mid \mathbf{rel})$ for each of the four instantiations **rel** of $\{V, M\}$. In our parameterization of the model, it turns out that this probability is $\geq 0.8$ only when **rel** assigns *false* to $V$ and *true* to $M$. So the CPD for $\eta[\varphi]$ specifies *true* with probability 1 in this case, and *false* with probability 1 in the other three cases. The resulting BN is illustrated in Figure 2.

In proving the correctness of this algorithm, we will use the following lemma:

**Lemma 1** *Let $M$ be a PEL model, $a \in \mathcal{A}$, $i \in \{1, \ldots, N_a\}$, and $s \in S$. Let $o_{a,i,s}$ be the instantiation of $\mathcal{O}_{a,i}$ that $s$ satisfies. Then for any formulas $\varphi$ and $\psi$:*

$$p_{a,i,s}([\varphi]_M \mid [\psi]_M)$$
$$= \mathcal{P}(\eta[\varphi] = true \mid \langle o_{a,i,s}; \eta[\psi] = true \rangle)$$

This lemma puts the criterion for satisfaction of $BelCond_{a,i}^{\geq r}(\varphi \mid \psi)$ in a more convenient form. The proof, which we do not give here, uses the definition of $p_{a,i,s}$ and the Observation Set assumption.

**Proposition 1 (Correctness of CreateNode)**
*Suppose an augmented BN $\mathcal{B}$ is $\Delta$-equivalent to a PEL model $M$. Then when **CreateNode**$(\mathcal{B}, \varphi)$ terminates, $\mathcal{B}$ is $(\Delta \cup \{\varphi\})$-equivalent to $M$.*

**Proof:** We use structural induction on $\varphi$; the inductive hypothesis is that Proposition 1 holds for all subformulas of $\psi$. Thus, the recursive calls at the beginning of **CreateNode** make it so $\mathcal{B}$ is $\Gamma$-equivalent to $M$, where $\Gamma$ is $\Delta$ plus all the subformulas of $\varphi$. Then **CreateNode**$(\mathcal{B}, \varphi)$ adds $\eta[\varphi]$ to $\mathcal{B}$. Let Pr be the distribution defined by $\mathcal{B}$ before this addition, and Pr' be the distribution afterwards.

By the definition of $\Gamma$-equivalence, we know that for every instantiation $\mathbf{x}$ of $\pi(\Phi) \cup \eta[\Gamma]$, $\mathcal{P}(\mathbf{x}) = \Pr(\mathbf{x})$. We must show that for every instantiation $\langle \mathbf{x}; (\eta[\varphi] = t) \rangle$ of $\pi(\Phi) \cup \eta[\Delta] \cup \{\eta[\varphi]\}$:

$$\mathcal{P}(\langle \mathbf{x}; \eta[\varphi] = t \rangle) = \Pr'(\langle \mathbf{x}; \eta[\varphi] = t \rangle) \quad (1)$$

Let **pa** be the instantiation **x** limited to the parents of the newly created node $\eta[\varphi]$. Then by the definition of conditional probability and the chain rule for Bayes nets, equation (1) is equivalent to: $\mathcal{P}(\eta\,[\varphi] = t \mid \mathbf{x}) \cdot \mathcal{P}(\mathbf{x}) = \Pr'(\eta\,[\varphi] = t \mid \mathbf{pa}) \cdot \Pr'(\mathbf{x})$. **CreateNode** only adds nodes as children of existing nodes, so the marginal distribution over existing nodes is not changed. Thus $\Pr'(\mathbf{x}) = \Pr(\mathbf{x})$. Along with the fact that $\mathcal{P}(\mathbf{x}) = \Pr(\mathbf{x})$, this allows us to reduce equation (1) to:

$$\mathcal{P}(\eta\,[\varphi] = t \mid \mathbf{x}) = \Pr'(\eta\,[\varphi] = t \mid \mathbf{pa}) \qquad (2)$$

So what we must show is that the CPDs defined by **CreateNode** are the correct CPDs for the indicator variables. The cases where $\varphi$ is an atomic or Boolean formula are straightforward, so we move directly to the interesting case where $\varphi = BelCond_{a,i}^{\geq r}(\psi \mid \chi)$.

In this case, **CreateNode** adds a node $\eta\,[\varphi]$ with the relevant members of $\mathcal{O}_{a,i}$ as parents (to simplify the presentation, we will assume that all members of $\mathcal{O}_{a,i}$ are relevant). Suppose $\mathbf{x}$ assigns values $\mathbf{o}_{a,i}$ to $\mathcal{O}_{a,i}$. Then what we have to prove is:

$$\mathcal{P}(\eta\,[\varphi] = t \mid \mathbf{x}) = \Pr'(\eta\,[\varphi] = t \mid \mathbf{o}_{a,i}) \qquad (3)$$

Consider any state $s$ in which $\mathbf{x}$ holds. By Lemma 1 and the definition of satisfaction, $\eta\,[\varphi]\,(s) = true$ if and only if: $\mathcal{P}(\eta\,[\psi] = true \mid \mathbf{o}_{a,i,s}, \eta\,[\chi] = true) \geq r$. But $\psi$ and $\chi$ are subformulas of $\varphi$, and $\mathcal{O}_{a,i} \subset \Phi$, so by the assumption that $\mathcal{B}$ is $\Gamma$-equivalent to $M$:

$$\mathcal{P}(\eta\,[\psi] = true \mid \mathbf{o}_{a,i}, \eta\,[\chi] = true)$$
$$= \Pr(\eta\,[\psi] = true \mid \mathbf{o}_{a,i}, \eta\,[\chi] = true)$$

This last probability value is exactly what **CreateNode** compares to $r$ in constructing the CPD for $\eta\,[\varphi]$. So the CPD is correct. ∎

Once we have a BN that is $\Delta$-equivalent to $M$ we can assert any formula $\varphi \in \Delta$ by setting $\eta\,[\varphi] = true$ as evidence. To find the probability of any formula $\varphi \in \Delta$, we simply query $\eta\,[\varphi] = true$. For example, the formula $\varphi = Bel_{i,4}^{\geq 0.8}((C = h) \vee (C = m))$ that we discussed earlier has probability 0.16 in our model. However, if we assert $V = false$ (i.e., the vaccine is ineffective), then $\Pr(\varphi)$ goes up to 0.8.

The number of BN queries required to make a BN $\Delta$-equivalent to $M$ is linear in the number of *BelCond* formulas, since **CreateNode** is only called once for each subformula. The **CreateNode** function takes time exponential in the maximal number of relevant observations for the *BelCond* subformulas, as we need to compute the probability $\Pr(\eta\,[\psi] \mid \langle \mathbf{rel}; \eta\,[\chi] = true \rangle)$ for every instantiation **rel** of **Rel**. Most naively, we simply run BN inference for each **rel** separately. In certain cases, we can get improved performance by running a single query $\Pr(\eta\,[\psi], \mathbf{Rel} \mid \eta\,[\chi] = true)$ and then renormalizing appropriately; this approach can allow us to exploit the dynamic programming of BN inference algorithms. We note that the newly added nodes also add complexity to the BN, and can make the inference cost grow in later parts of the computation (e.g., by increasing the size of cliques).

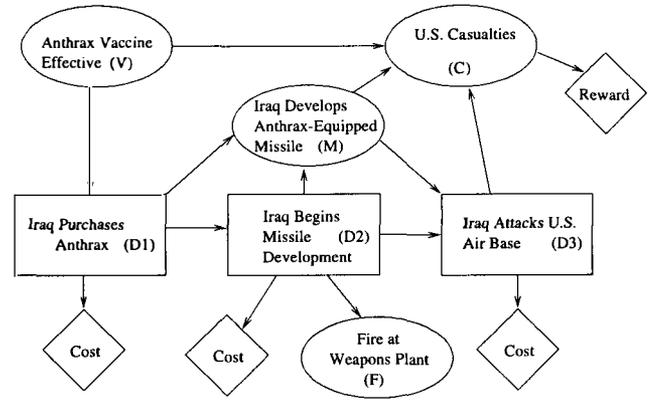

Figure 3: Influence diagram representing Iraq's decision scenario. No-forgetting arcs are not shown

## 5 Reasoning about Decisions

So far, we have assumed that we have a probability distribution over all variables in the system. In practice, however, we have agents who make decisions in accordance with their beliefs and preferences. In our example, $P$, $B$ and $A$ are actually decisions made by Iraq. Our construction took these to be random variables, each with a CPD representing a distribution over Iraq's decision. If these CPDs are reasonable, then our system will give reasonable answers; e.g., we will obtain a lower probability that the anthrax vaccine is effective if we observe that Iraq has purchased anthrax, since it would not be rational for Iraq to purchase a bacterium for which the U.S. has an effective vaccine. We would like to extend our framework to induce automatically the actions that agents will take at various decision points. As discussed in the introduction, this problem is quite complex when there are multiple decision makers with conflicting goals. We therefore focus on the case of a single decision maker. We note, however, that we can still have multiple agents reasoning about the decision maker and about each other's state of knowledge.

Assuming that agents act rationally, we can automate the construction of CPDs for decision nodes by modeling the decision maker's decision process with an influence diagram, and solving the influence diagram to obtain CPDs for the decision nodes. Somewhat surprisingly, the possibility of modeling other agents with influence diagrams has not been explored deeply in the existing literature, although Nilsson and Jensen mention it in passing [11]. Suryadi and Gmytrasiewicz [15] take an approach similar to ours in that they use an ID to model another agent's decision process. However, they discuss learning the structure and parameters of the ID from observations collected over a large set of similar decision situations. We assume that the ID is given, and concentrate on the inferences that can be made only from a few observations about the current situation.

Figure 3 depicts an influence diagram for the scenario described in the introduction. An influence diagram is a directed acyclic graph with three kinds of nodes. Chance nodes, like nodes in a BN, correspond to random variables; they are represented by ovals. Decision nodes, drawn as rectangles, correspond to variables that the decision-maker can control. Utility nodes, drawn as diamonds, correspond to components of the decision-maker's utility function.

The decision nodes of an ID are ordered $D_1, \ldots, D_n$ according to the order in which the decisions are made. The parents of $D_i$, denoted $Pa(D_i)$, are those variables whose value the decision-maker knows when decision $D_i$ is made. Thus, when we are creating a PEL model and an ID for the same scenario, the decision-maker's observation stages correspond to his decision nodes, with $\mathcal{O}_{a,i}$ equal to $Pa(D_i)$. A utility node $U_i$ represents a deterministic function $f_i$ from instantiations of $Pa(U_i)$ to real numbers. The utility of an outcome is the sum of the individual utility functions $f_i$.

Solving an influence diagram means deriving an optimal policy, consisting of a decision rule for each decision node. A decision rule $\delta_i$ for a node $D_i$ is a function from $dom(Pa(D_i))$ to $dom(D_i)$. For each instantiation of $Pa(D_i)$, the decision rule gives the action that maximizes the decision-maker's expected utility, assuming it will act rationally in all future decisions. The standard algorithms for solving IDs utilize backwards induction: the decision rules for the decision nodes are calculated in the reverse of their temporal order [9].

After we have the decision rules, we can easily transform an influence diagram $\mathcal{D}$ into a Bayesian network $\mathcal{B}(\mathcal{D})$. We remove the utility nodes, and change the decision nodes into chance nodes (ordinary BN nodes). If $D_i$ is a decision node, then for each instantiation **pa** of $Pa(D_i)$, we create a probability distribution that gives probability 1 to $\delta_i(\mathbf{pa})$, and probability 0 to all other elements of $dom(D_i)$. This distribution becomes the CPD for $D_i$ given **pa**.

We can use this system to make inferences about unobserved world variables based on evidence of agents' actions. Suppose the parameters of the model $\mathcal{D}$ depicted in Figure 3 are such that the prior probability of the vaccine being effective is 0.8, but it is irrational for Iraq to purchase anthrax unless it has observed the vaccine to be ineffective. As above, we may need to add some additional nodes to $\mathcal{D}$, such as *Observed* and *ObservedValue* nodes to model the U.S.'s uncertainty about whether Iraq observes $V$ at stage 1. We then use the method described in this section to derive CPDs for Iraq's decision nodes, creating a BN $\mathcal{B}(\mathcal{D})$. The influence diagram defines the observation sets for Iraq; we will use the U.S. observation sets described in Section 2. We can then use the algorithm of Section 4 to find the probabilities of arbitrary PEL formulas in the PEL model corresponding to $\mathcal{B}(\mathcal{D})$.

At stage 1, the U.S. assigns probability 0.8 to the vaccine being effective: all states satisfy $Bel_{u,1}^{\geq 0.8}(V = true)$. At stage 2, however, the situation changes. It turns out that $Bel_{u,2}^{\geq 0.8}(V = true)$ is true if and only if there is not a fire at the Iraqi biological weapons plant. A fire provides the U.S. with strong evidence that Iraq has begun developing an anthrax-carrying missile, which would not be rational unless Iraq had purchased anthrax, which implies that Iraq has observed the anthrax vaccine to be ineffective. So in this model, $\Pr(Bel_{u,2}^{\geq 0.8}(V = true)) = \Pr(F = false)$. In a more complex query, we could compute $\Pr(Bel_{u,2}^{\geq 0.8}(V = true) \mid V = false)$, i.e., the probability that the U.S. will believe the vaccine to be effective despite the fact that it is not. The answer to this query would depend on the prior probability about the vaccine's effectiveness, Iraq's decisions, and the chances of observing a fire.

The CPDs for decision nodes derived by solving an influence diagram become part of the common prior distribution in the resulting BN. However, these CPDs are derived using the decision-maker's utility function. Thus, in assuming that the decision-maker's decision rules are part of the common prior, we are implicitly assuming that the decision-maker's utility function is common knowledge. Like the assumption that observations are common knowledge, this is an assumption we would like to relax.

Just as we introduced *Observes* nodes to model uncertainty about an agent's observations, we can introduce preference nodes to model uncertainty about an agent's utility function. These preference nodes are parents of particular utility nodes, and modify the way the utility depends on other variables. They are also in all the decision-maker's observation sets, assuming he knows his own preferences. One might propose to use continuous-valued preference nodes that define a distribution over the decision-maker's utility value. The problem with this approach is that these continuous-valued preference nodes must be parents of every decision node, and standard ID solution algorithms cannot handle continuous values in such a context. We therefore use discrete-valued preference nodes, with the resulting coarse-grained preference models. For example, we can introduce a node $A$ representing Iraq's aversion to doing business with criminal biological weapons dealers, which is a parent of the cost node associated with $D_1$. If $A = high$, then the cost is greater in magnitude than it would be if $A = low$. The preference node $A$ is in the parent sets of all Iraq's decision nodes, but the U.S. will not be able to observe it directly.

## 6  Discussion and Future Work

This paper combines epistemic logic with Bayesian networks to create an integrated system for probabilistic reasoning about agents' beliefs and decisions. Although PEL is essentially a restricted version of the logic presented by Fagin and Halpern, we believe it is flexible enough to be useful in many practical applications. Furthermore, the simplicity of PEL allows us to define an algorithm for finding the probability of a formula in a PEL model using a Bayesian network, rather than constructing the PEL model explicitly. We also show how to construct this Bayesian network from an influence diagram, rather than having a human fill in the CPDs for nodes that represent an agent's decisions.

Our approach is limited by the common prior assumption, which implies that all differences between agent's beliefs are due to their having different observations. This assumption is common in economics, and has important ramifications [1]. It allows agents' beliefs to be arbitrarily different, as long as they have received sufficiently different observations. But it may be impractical to represent in a BN all the different observations that have caused agents' beliefs to diverge. An alternative is to explicitly represent uncertainty about each agent's probability distribution. However, this approach introduces substantial complications: Do we also model one agent's distribution about another agent's distribution? If so, do we model the infinite belief hierarchy? Therefore, the extension to this case is far from obvious. Another assumption that we would like to relax is that agents are perfect probabilistic reasoners and decision makers.

The other obvious limitation of the system described in this paper is that although it can reason about the beliefs of an arbitrary number of agents, it can only reason explicitly about one agent's decisions. If we wish to have the system automatically derive the CPDs for decisions made by multiple agents, the maximum expected utility solution concept is no longer appropriate, since the agents do not have probability distributions over each other's actions. We could utilize game-theoretic solution concepts [5] to find rational strategies for the agents, and then substitute these strategies for the agents' CPDs as we did in Section 5; the rest of our results would still be applicable. However, the framework of multi-agent rationality is substantially more ambiguous than the single agent case, so that this approach does not define a unique answer. We hope to investigate this issue in future work.

**Acknowledgments.** We thank Yoav Shoham for useful discussions and Uri Lerner and Lise Getoor for their work on the PHROG system. This work was supported by ONR contract N66001-97-C-8554 under DARPA's HPKB program.